\documentclass[10pt,twocolumn,letterpaper]{article}

\usepackage{iccv}
\usepackage{times}
\usepackage{epsfig}
\usepackage{graphicx}
\usepackage{amsmath}

\usepackage{amssymb,pifont}
\newcommand{\cmark}{\ding{51}}%
\newcommand{\xmark}{\ding{55}}%
\usepackage{booktabs}
\usepackage{colortbl}
\usepackage{bigstrut}
\usepackage{multirow}

\usepackage{cite}
\usepackage{makecell}
\usepackage{boldline,bm}
\setcellgapes{3pt}
\usepackage{arydshln}
\usepackage{tabu}
\usepackage{subcaption}
\usepackage{balance}
\usepackage[marginal]{footmisc}

\usepackage[pagebackref=true,breaklinks=true,letterpaper=true,colorlinks,bookmarks=false]{hyperref}

\newcommand{\bigsize}{\fontsize{9.8pt}{10pt}\selectfont}
\usepackage[breaklinks=true,bookmarks=false]{hyperref}

\iccvfinalcopy % *** Uncomment this line for the final submission

\def\iccvPaperID{****} % *** Enter the ICCV Paper ID here

\ificcvfinal\fi
\begin{document}

%%%%%%%%% TITLE
\title{RANet: Ranking Attention Network for Fast Video Object Segmentation}

\author{Ziqin Wang$^{1,3}$, Jun Xu$^{2,4}$\thanks{Corresponding author: Jun Xu (nankaimathxujun@gmail.com).\ This work is done when Ziqin Wang was an intern in IIAI.}, Li Liu$^{2}$, Fan Zhu$^{2}$, Ling Shao$^{2}$\\
$^1$The University of Sydney, Sydney, Australia\\
$^2$Inception Institute of Artificial Intelligence (IIAI), Abu Dhabi, UAE\\
$^3$Institute of Artificial Intelligence and Robotics, Xi'an Jiaotong University, Xi'an, China\\
$^4$Media Computing Lab, College of Computer Science, Nankai University, Tianjin, China\\
{\tt\small Project page:\ https://github.com/Storife/RANet}
}

%Ziqin Wang (Xi'an Jiaotong University) <782908504@qq.com>
%Jun Xu (Inception Institute of Artificial Intelligence) <nankaimathxujun@gmail.com>
%Li Liu (the inception institute of artificial intelligence) <liuli1213@gmail.com>
%Fan Zhu (Inception Institute of Artificial Intelligence) <fanzhu1987@gmail.com>
%Ling Shao (Inception Institute of Artificial Intelligence) <ling.shao@ieee.org>

%\and

\maketitle
\thispagestyle{empty}
\pagestyle{empty}

%%--------ABSTRACT--------%% 
\begin{abstract}
Despite online learning (OL) techniques have boosted the performance of semi-supervised video object segmentation (VOS) methods, the huge time costs of OL greatly restrict their practicality.\ Matching based and propagation based methods run at a faster speed by avoiding OL techniques.\ However, they are limited by sub-optimal accuracy, due to mismatching and drifting problems.\ In this paper, we develop a real-time yet very accurate Ranking Attention Network (RANet) for VOS.\ Specifically, to integrate the insights of matching based and propagation based methods, we employ an encoder-decoder framework to learn pixel-level similarity and segmentation in an end-to-end manner.\ To better utilize the similarity maps, we propose a novel ranking attention module, which automatically ranks and selects these maps for fine-grained VOS performance.\ Experiments on DAVIS$_{16}$ and DAVIS$_{17}$ datasets show that our RANet achieves the best speed-accuracy trade-off, e.g., with $33$ milliseconds per frame and $\mathcal{J}\&\mathcal{F}$$=$$85.5\%$ on DAVIS$_{16}$.\ With OL, our RANet reaches $\mathcal{J}\&\mathcal{F}$$=$$87.1\%$ on DAVIS$_{16}$, exceeding state-of-the-art VOS methods.\ The code can be found at \url{https://github.com/Storife/RANet}.
%Moreover, our method can still achieve good performance when trained only on static images.
\end{abstract}

%%--------Introduction--------%% 
\vspace{-2mm}
\section{Introduction}
\begin{figure}[t]
\vspace{-2mm}
\begin{center}
\includegraphics[width=1.0\linewidth]{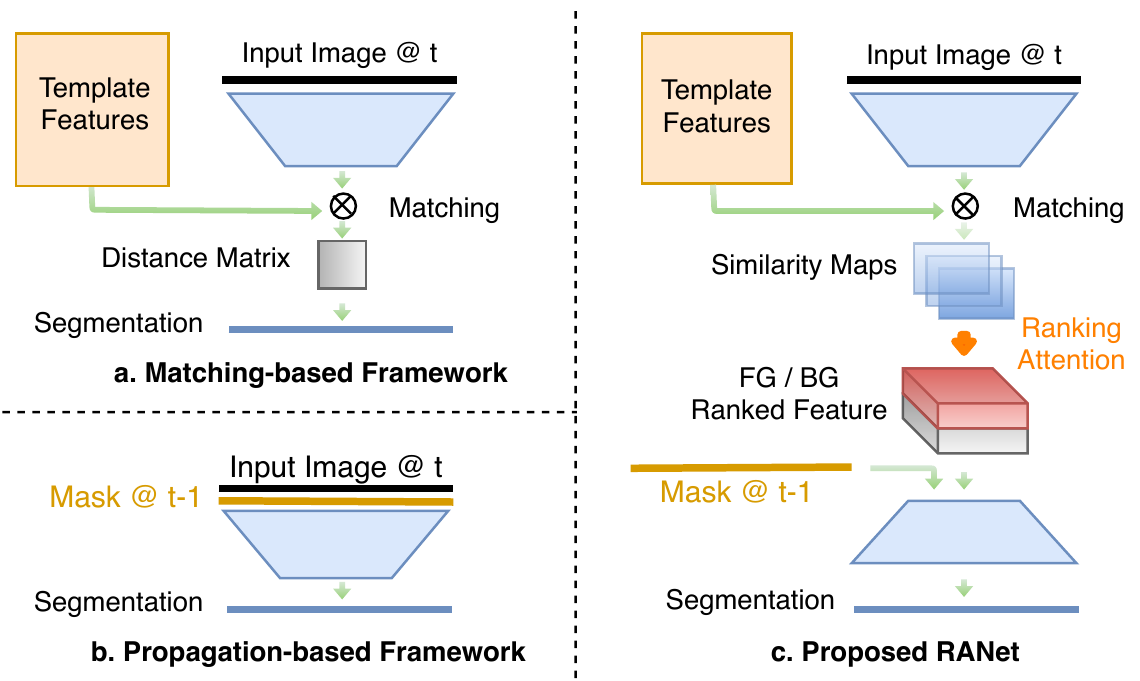}
\end{center}
\vspace{-6mm}
\caption{\textbf{Comparison of different VOS frameworks}.\ (\textbf{a}) Matching based framework;\ (\textbf{b}) Propagation based framework; and (\textbf{c}) Proposed RANet.\ We propose a novel \textsl{Ranking Attention} module to rank and select important features.}
\label{fig:intro}
\vspace{-6mm}
\end{figure}

Semi-supervised Video Object Segmentation (VOS)~\cite{davis2016,davis2017,davis2018} aims to segment the object(s) of interests from the background throughout a video, in which only the annotated segmentation mask of the first frame is provided as the template frame at test phase.\ This challenging task is of great importance for large scale video processing and editing~\cite{semi02,semi01,zsvos}, and many video analysis applications such as video understanding~\cite{Fan2019VideoSal,un02} and object tracking~\cite{siammask}.\ 

Early VOS methods~\cite{osvos, masktrack, osvos-s, onavos} mainly resort to online learning (OL) techniques which fine-tune a pre-trained classifier on its first frame.\ Matching or propagation based methods have also been proposed for VOS.\ Matching based methods~\cite{videomatch, pml} segment pixels according to the pixel-level matching scores between the features of the first frame and of each subsequent frame (Fig.~\ref{fig:intro}\ (\textbf{a})), while propagation based methods~\cite{masktrack, rgmp, favos, osmn, sfl, semi02} mainly rely on temporally deforming the annotated mask of the first frame via predictions of the previous frame~\cite{masktrack} (Fig.~\ref{fig:intro}\ (\textbf{b})).

% The respective benefits and drawbacks 
The respective benefits and drawbacks of these methods are clear.\ Specifically, OL based methods~\cite{osvos, masktrack, osvos-s, onavos} achieve accurate VOS at the expense of speed, requiring several seconds to segment each frame~\cite{osvos}.\ On the contrary, simple matching or propagation based methods~\cite{masktrack,pml,plm} are faster, but with sub-optimal VOS accuracy.\ Matching based methods~\cite{videomatch, pml, rgmp} bear up the mismatching problem, \ie, violating the temporal consistency of the primary object with constantly changing appearance in the video.\ On the other hand, propagation based methods~\cite{masktrack, rgmp, favos, osmn, sfl, ofl} suffer from the drifting problem due to occlusions or fast motions between two sequential frames.\ In summary, most existing methods cannot tackle the VOS task with both satisfactory accuracy and speed, which are essential for practical applications.\ More efficient methods are still required to reach a better speed-accuracy trade-off for the VOS task. 

With the above considerations, in this work, we develop a real-time network for fine-grained VOS performance.\ The developed network benefits from an encoder-decoder structure, and learns pixel-level matching, mask propagation, and segmentation in an end-to-end manner.\ Fig.~\ref{fig:intro} (\textbf{c}) shows a glimpse of the proposed network.\ A Siamese network~\cite{siamfc} is employed as the encoder to extract pixel-level matching features, and a pyramid-like decoder is used for simultaneous mask propagation and high-resolution segmentation.\ 
 
%dynamic numbers of foreground (FG) and background (BG) similarity maps produced by 

A key problem in our framework is how to connect the pixel-level matching encoder and propagation based decoder in a meaningful manner.\ The encoder produces dynamic foreground and background similarity maps, which cannot be directly fed into the decoder.\
%since the numbers of similarity maps for FG\ /\ BG varies with the number of the FG\ /\ BG pixels in template frame, and then 
%discrepancy between the regular convolutional layers of current networks and the irregular appearance and locations of the object(s) and background in videos.\ 
To this end, we propose a \emph{Ranking Attention Module} (RAM, see Fig.~\ref{fig:intro}\ (\textbf{c})) to reorganize (\ie, rank and select) the similarity maps according to their importance for fine-grained VOS performance.\ The proposed \emph{Ranking Attention Network} (RANet) can better utilize the pixel-level similarity maps for fine-grained VOS, greatly alleviating the drawbacks of previous matching or propagation based methods.\ Experiments on DAVIS$_{16}$ and DAVIS$_{17}$ datasets~\cite{davis2016,davis2017} demonstrate that the proposed RANet outperforms previous VOS methods in terms of speed and accuracy, e.g., achieving $\mathcal{J}\&\mathcal{F}=85.5\%$ at a speed of $30$ FPS on DAVIS$_{16}$.
% (frames per second) 
%Additionally, when armed with OL techniques, RANet outperforms all state-of-the-art methods on commonly tested VOS datasets. 

The contributions of this work are three-fold: 
\vspace{-2mm}
\begin{itemize}
\item
We integrate the benefits of matching and propagation frameworks in an end-to-end manner and develop a real-time network for the semi-supervised VOS task. 
\vspace{-3mm}
\item 
\vspace{-3mm}
We propose a novel \emph{Ranking Attention Module} to rank and select conformable feature maps according to their importance for fine-grained VOS performance.
\vspace{-2mm}
\item 
Experiments on DAVIS$_{16}/_{17}$ datasets show that the proposed RANet achieves competitive or even better performance than previous VOS methods, at real-time speed.\ The proposed RANet achieves accurate VOS results even been trained only with static images.
\end{itemize}
\vspace{-4mm}
%We largely push back the frontier of speed-accuracy trade-off in VOS task through an end-to-end network.

%%--------Related Works--------%% 
\section{Related Works}
\noindent
{\bf Online learning based methods}.\
OL based methods~\cite{osvos,onavos,masktrack,osvos-s,reid,lucid,premvos,premvos2,premvos3} fine-tune on the first frame of a video to extract the primary object(s), and then segment the video frame-by-frame.\ OSVOS~\cite{osvos} uses a pre-trained object segmentation network, and fine-tunes it on the first frame of the test video.\ OnAVOS~\cite{onavos} extends OSVOS with an online adaptation mechanism, and OSVOS-S~\cite{osvos-s} utilizes semantic information from an instance segmentation network.\ LucidTracker~\cite{lucid} introduces a data augmentation mechanism for online fine-tuning.\ DyeNet~\cite{reid} integrates instance re-identification and temporal propagation, and uses OL to boost the performance.\ PReMVOS~\cite{premvos,premvos2,premvos3} integrates techniques from instance segmentation~\cite{mask-rcnn}, optical flow~\cite{flownet,flownet2}, refinement, and re-identification~\cite{personsearch} together with extensive fine-tuning, and achieves satisfactory performance.\ In summary, OL is very effective for the VOS task.\ Therefore, subsequent methods~\cite{masktrack, cinm, reid} regard OL as a conventional technique to boost VOS performance.\ However, OL based methods are computationally expensive for practical applications.\ In this work, we solve the VOS problem with a very fast network that obtains a competitive accuracy at a speed of $30$ FPS on DAVIS$_{16}$, $130\sim400$ times faster than previous OL based methods~\cite{osvos, masktrack, osvos-s, onavos}.

%MaskTrack~\cite{masktrack} learns to propagate the segmentation mask from one frame
%to the next, and introduce OL to reduce the drifting problem of propagation. 

%Hu et al. [15] propose a motion-guided cascaded refinement network that uses a coarse segmentation obtained by an active contour model on optical flow as guidance.
%MaskRNN [16] uses recurrent neural networks to fuse
%the output of two deep networks. LSE [9] proposes using a location-sensitive embedding to refine an initial foreground
%prediction. MoNet [39] exploits optical flow motion cues by feature alignment and a distance transform layer.
%Han et al. [13] propose a cutting-policy network which is
%trained with reinforcement learning to estimate a region of
%interest which is segmented by a cutting-execution network.

%This strategy is very effective and achieve high performance on VOS task, but meanwhile brings huge computational burden to these methods, hindering them from practical scenarios.\ 

\begin{figure*}[t]
\vspace{-5mm}
\begin{center}
\includegraphics[width=1\linewidth]{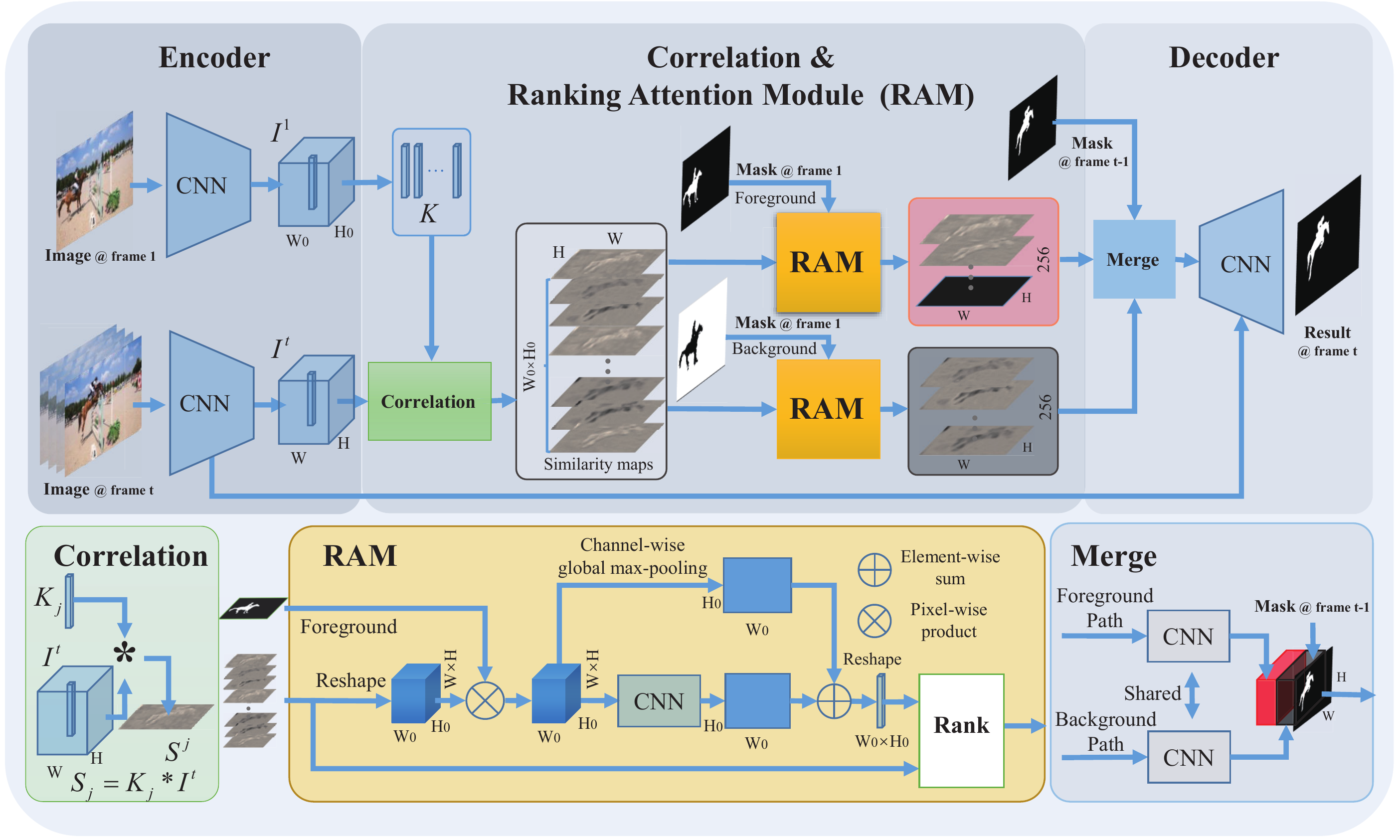}
\end{center}
\vspace{-7mm}
\caption{\textbf{Illustration of the proposed RANet}.\ We compute correlation of the features extracted by Siamese networks.\ The output similarity maps and template mask are fed into the RAM module to rank and select the foreground/background similarity maps.\ Then these maps and the previous frame's mask and fed into the decoder for final segmentation.}
\label{fig:our}
\vspace{-5mm}
\end{figure*}

\noindent
{\bf Propagation or matching based methods}.\ Propagation based methods additionally resort to the previous frame(s) for better VOS performance.\ Masktrack\cite{masktrack} tackles VOS by combining the image and segmentation mask of the previous frame as the input.\ This strategy is also used in CINM~\cite{cinm}, OSMN~\cite{osmn} and RGMP~\cite{rgmp}.\ 
%Since simple propagation strategy tend to cause the drifting problem, 
%Masktrack introduces OL to remember the target object. 
%The architecture of RGMP is the most similar one to our RANet, but in our RANet, we use pixel-level matching technique instead of simply stacking, and we feed the previous frame's mask into the decoder, instead of the encoder as RGMP does.\ 
RGMP~\cite{rgmp} stacks the first, previous and current frames' features during propagation through a Siamese architecture network.\ In this work, we also utilize the Siamese network, but use a pixel-level matching technique instead of simply stacking, and feed the previous frame's mask into the decoder, instead of the encoder as in RGMP~\cite{rgmp}.\ OSMN~\cite{osmn} introduces a modulator to manipulate the intermediate layers of the segmentation network, by using visual and spatial guidance.\
%RGMP~\cute{rgmp} tackles the VOS task via a Siamese architecture network.\ This leads to a faster speed when dealing with multi-instance videos since the features of the encoder can be shared among all instances.\ 
Optical flow~\cite{flownet,flownet2} is also used to guide the propagation process in many methods~\cite{sfl,masktrack,ofl,ctn}.\ However, it fails to distinguish non-rigid objects from motionless sections of the background.\ %, and requires extra running time for flow estimation. 
All these strategies are effective, but still, suffer from the drifting problem.\ MaskTrack~\cite{masktrack} embraces OL to remember the target object, which eliminates this problem and improves VOS performance.\ However, since OL is time-consuming, we employ more efficient matching techniques to handle this drifting problem.
%\noindent
%{\bf Matching based methods}.\
%Similar to online fine-tuning, Matching based methods~\cite{videomatch, pml, plm} do not rely on the temporal information, but 

Matching based methods~\cite{videomatch, pml, plm, Voigtlaender2019FEELVOS} are very efficient.\ They first calculate pixel-level matching between the features of the template frame and the current frame in videos, and then segment each pixel of the current frame directly from the matching results.\ Pixel-Wise Metric Learning~\cite{pml} predicts each pixel by nearest neighbor matching in pixel space to the template frame.\ However, the point-to-point correspondence strategy~\cite{deepmatching, plm} often results in noisy predictions.\ To ease this problem, we apply a decoder to utilize the matching results as guidance.\ Hu~\etal proposed a soft matching mechanism in VideoMatch~\cite{videomatch}, which performs soft segmentation upon the averaged similarity score maps of matching features to generate smooth predictions.\ However, due to the lack of temporal information, they still suffer from the mismatching problem.\ In this work, we employ both the strategies of point-to-point correspondence matching for pixel-level object location and temporal propagation, to handle the mismatching and drifting problem.\ 
FEELVOS~\cite{Voigtlaender2019FEELVOS} employs global and local matching for more stable pixel-level matching, but only calculates extreme value maps for final segmentation, losing major information of the similarity maps.\ Our RAM can better utilize the similarity information.\ Moreover, for faster speed, we use a light-weight decoder and employ a standard ResNet~\cite{resnet} pre-trained on ImageNet~\cite{imagenet} as the backbone, instead of the time-consuming semantic segmentation networks~\cite{deeplab, deeplabv3, deeplabv3+, YazhaoSegICCV2019} used in previous methods~\cite{videomatch, masktrack}.

\section{Proposed Method}
\label{sec:method}
In this section, we first provide an overview of the developed Ranking Attention Network (RANet) in~\S\ref{sec:networkoverview}.\ In~\S\ref{sec:ccsim}, we describe the proposed Ranking Attention Module (RAM), and extend it for multi-object VOS in~\S\ref{sec:mifg}.\ Finally, we present the implementation details and training strategies for RANet in~\S\ref{sec:imple} and~\S\ref{sec:train}, respectively.

%%--------Overview------%%
\subsection{Network Overview}
\label{sec:networkoverview}
Our RANet consists of three seamless parts:\ an encoder for feature extraction, an integration of correlation and RAM, and a decoder for feature merging and final segmentation.\ An illustration of our RANet is shown in Fig.~\ref{fig:our}.

%The cross-correlation is applied between filters and current frame features and produces a set of response maps. Note that the number of maps and filters are the same. Then, the response maps are fed into a feature merging network to process the foreground and background features. Finally, the combined features pass through the pyramid networks to obtain final results. Note that the prediction of a previous frame is concatenated into the combined features as time-sequential information, which was first done in Mask track~\cite{masktrack} and later used in RGMP~\cite{rgmp}. However, we place the previous frame masks before the decoder instead of using a 4-channel (RGB and a mask) input image.
%hence we can regard it as the reference frame 
%, where $C$ is the number of feature channels, $H_0/W_0$ and $H/W$ are the height/width of template and target feature maps, respectively.\ 
% .\ We utilize the pixel-level matching technique~\cite{pml} to locate the foreground (FG) object in the target, and separate it from the background (BG).\ To avoid the drifting problem, we match the target features using the corresponding template features, since the template has the annotated segmentation mask in semi-supervised VOS task.
%rather than the previous frame as in~\cite{masktrack, rgmp, favos, osmn, sfl, ofl}
% Denote  and  as the template and target feature tensors extracted by the Siamese encoder

% The purpose of the correlation layer is to calculate the similarity between the template features $\bm{I}^{1}$ and the target features $\bm{I}^{t}$.

\noindent
\textbf{Siamese Encoder}.\ 
To obtain correlation information for accurate VOS, we employ Siamese networks~\cite{siamfc} (with shared weights) as the encoder to extract features of the first frame and the current frame.\
Then we extract pixel-level features from the first frame, reshape it into a conformable shape, as the template features for correlation calculation.

\noindent
\textbf{Correlation and RAM for Matching}.\
Correlation is widely used in object tracking.\ In SiamFC~\cite{siamfc}, correlation is used to locate the position of the object using similarity maps.\ In our RANet, to locate each pixel of the object(s) for segmentation, we need pixel-level similarity maps by calculating the correlation between each pixel-level feature of the template and current frames.\ Note that there is one similarity map for each pixel-level template feature.\ The detailed formulation of correlation will be described in \S\ref{sec:ccsim}.\ We then utilize the mask of the first frame to select foreground (FG) or background (BG) similarity maps as FG or BG features for segmentation.\ Since the number of FG or BG pixels varies in different videos, the number of FG or BG similarity maps is dynamic, and hence the decoder has to deal with FG or BG similarity features with dynamic channel sizes.\ To handle this dynamic channel-size problem, we propose a RAM module to rank and select the most important similarity maps and organize them in conformable shape.\ This part will also be exhaustively explained in \S\ref{sec:ccsim}.\
%The correlation layer computes the pixel-level matching similarity maps between the template and current frame's feature,
The RAM module provides abundant and ordered features for segmentation, and leads to better performance, as will be shown in the ablation study in \S\ref{sec:ablation}.\ For simplicity, here we only consider the single-object VOS in \S\ref{sec:ccsim}.\ Extension of our RANet for multi-object VOS will be described in \S\ref{sec:mifg}.\ 
%To accomplish this, the ground truth mask of the reference frame is used to separate the foreground and background reference features for cross-correlation.
%Notice that we add a pooling layer after the reference stream encoder to improve robustness and reduce calculations by decreasing the number of filters. Thus, the size of $\bm{I}^{1}$ is half the size of $I^t$.
%Inspired by the SiamFC network~\cite{siamfc}, we use a correlation layer to locate each pixel in the template frame by measuring the pixel-level similarity maps between the template features and current frame features.\ 

\noindent
\textbf{Propagation}.\ 
Here we utilize the simple mask propagation method~\cite{masktrack}, while other propagation~\cite{reid, flownet2} or local-matching~\cite{Voigtlaender2019FEELVOS} methods would potentially improve our RANet.\ 
%The outcomes of RAM are two sets of ranked feature maps describing the FG and BG similarity between the template and current frame, respectively.\ 
We feed the predicted mask of the previous frame, together with the selected features of FG (or BG) by the proposed RAM, into the subsequent decoder.\ In this way, our RANet utilizes both matching and propagation techniques.

\noindent
\textbf{Light-weight Decoder}.\
This part contains a merge module and a pyramid-like network, which are described in the \textit{Supplementary File}.\ The merge module refines the two streams of ranked similarity maps, and then concatenates these maps with previous frame's mask.\ In the merge module, the two streams of the network share the same parameters.\ A pyramid-like network~\cite{u-net,refinenet,accv} is employed to obtain the final segmentation, with skip-connections to utilize multi-scale features of different layers.\ 
% Note that we add the previous frame's mask before the decoder.\
%multi-scale skip-connection

%Each level has a Multi-scale block, two Conv-block and a Res-block as is shown in Fig.~\ref{fig:merge}. 
%The Multi-scale block has three branches of convolutional layers with dilated sizes of 1, 6, and 9. All the branches are merged through element-wise summation.

%%--------CRAM--------%% 
\subsection{Correlation and Ranking Attention Module}
\label{sec:ccsim}
%Correlation is widely used in object tracking. In SiamFC, correlation is used to locate the position of the whole object using the produced similarity map. In our RANet, to locate each pixel of the object(s) for segmentation, we need pixel-level similarity maps by calculating the correlation between features
%of each pixel in the template frame and that in the current frame. 
%The correlation layer computes the pixel-level matching similarity maps between the template and current frame's feature,
%To provide abundant and ordered features for segmentation, the RAM is proposed to automatically rank and select the similarity maps.\ For simplicity, here we only consider the single object VOS problem.\ Extension to multi-instance VOS will be discussed in \S\ref{sec:mifg}.\ 

%To accomplish this, the ground truth mask of the reference frame is used to separate the foreground and background reference features for cross-correlation.
%Notice that we add a pooling layer after the reference stream encoder to improve robustness and reduce calculations by decreasing the number of filters. Thus, the size of $\bm{I}^{1}$ is half the size of $I^t$.
\begin{figure}
\vspace{-5mm}
\begin{center}
\includegraphics[width=1\linewidth]{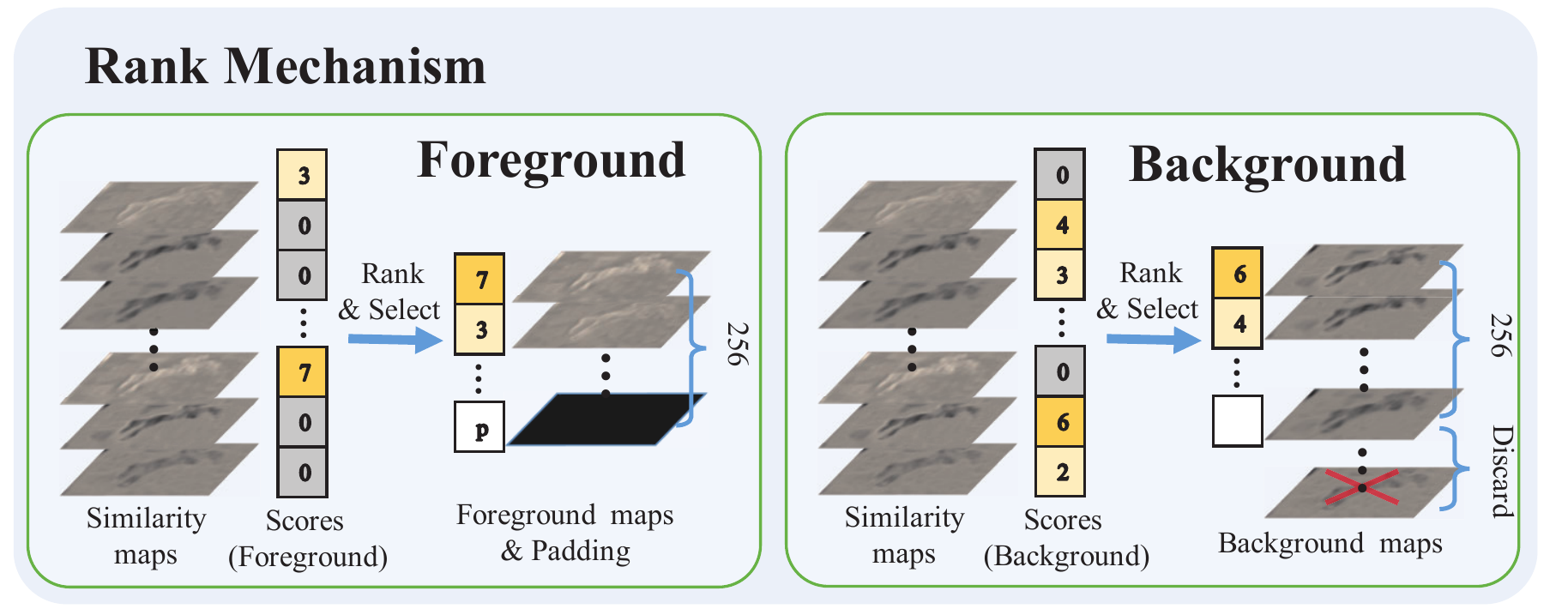}
\end{center}
\vspace{-7mm}
\caption{\textbf{Mechanism of the proposed Ranking Attention Module}.\ In FG (or BG) path, only the FG (or BG) similarity maps are selected.\ The maps are ranked from top to bottom according to ranking scores learned from attention network, and padding or discarding is operated to make the $256$ FG (or BG) maps.\ Finally, these maps are concatenated across the channel as features with the size of $256 \times H \times W$.}
\label{fig:rank}
\vspace{-4mm}
\end{figure}

\noindent
\textbf{Correlation}.\ We utilize correlation to find matching between pixels in the template and current frames.\ Denote $\bm{I}^{1}\in\mathbb{R}^{C\times H_0\times W_0}$ and $\bm{I}^{t}\in\mathbb{R}^{C\times H\times W}$ as the feature of template and current frames, extracted by the Siamese encoder, where $C$ is the number of feature channels, Denote $H_0$ ($W_0$) and $H$ ($W$) as the height (width) of template and current frame feature maps, respectively.\ 
%The purpose of the correlation layer is to calculate the similarity between the template features $\bm{I}^{1}$ and the current frame features $\bm{I}^{t}$.
We reshape the template features $\bm{I}^{1}\in\mathbb{R}^{C\times H_0\times W_0}$ to the size of $H_0W_0 \times (C \times 1 \times 1)$.\ Denote the reshaped template feature set as $\mathcal{K}=\{K_j|j=1,..,H_0\times W_0\}$, which consist of $H_0\times W_0$ features with the size of $C \times 1 \times 1$.\ 
%!!!According to the template mask, we assume that there are $N_{1}$ foreground and $N_{0}$ background pixels in template frame (pixels in feature space $\mathbb{R}^{C\times H_0\times W_0}$), then we have $N_{1}+N_{0}=H_0\times W_0$. 
%!!!, and then divide it into two set of features, object template feature set $\mathcal{K}^{1}=\{K^1_j|j=1,..,N_1\}$ and background template feature set $\mathcal{K}^{0}=\{K^0_j|j=1,..,N_0\}$, which consist of $N_{1}$ and $N_{0}$ features with the size of $C \times 1 \times 1$, respectively. The combination of these two set is noted as $\mathcal{K}=\{K^i_j|i=0,1; j=1,..,H_0\times W_0\}$. 
%****************************************************************
%The template features $\bm{I}^{1}$ contain $H_0 \times W_0$ pixels in feature space, and we extract all the pixel-level features (of size $C\times1\times1$) and denote them as a set $\mathcal{K}$.\ Note that the number of elements in $\mathcal{K}$ is $H_0\times W_0$.\ 
%The template features $\bm{I}^{1}$ contain $H_0 \times W_0$ pixels in feature space, and we extract all the pixel-level features (of size $C\times1\times1$) and denote them as a set $\mathcal{K}$.\ Note that the number of elements in $\mathcal{K}$ is $H_0\times W_0$.\ 
%We divide $\mathcal{K}$ into the object feature set $\mathcal{K}^{1}$ and background feature set $\mathcal{K}^{0}$, according to the template mask of size $H_0\times W_0$.\ Denote the number of elements in $\mathcal{K}^{1}$ and $\mathcal{K}^{0}$ as $N_{1}$ and $N_{0}$, where $N_{1}+N_{0}=H_0\times W_0$.\ 
In our RANet, the correlation is computed between the $\ell_2$-normalized features $K_j$ in template frame $\mathcal{K}$ and the current frame $\bm{I}^{t}$.\ After correlation, we have the similarity maps $\bm{S}_j = \bm{K}_j \ast \bm{I}^t$ whose size is $W\times H$.\
%We denote $\mathcal{S}^1$ ($\mathcal{S}^0$) as the set of object (background) similarity maps.
Denote the tensor $\mathcal{S}\in\mathbb{R}^{H_0W_0\times H\times W}$ as the set of correlation maps.\ Then we have
\begin{equation}
\mathcal{S} = \{\bm{S}_j\ |\  \bm{S}_j = \bm{K}_j \ast \bm{I}^t\}_{
%i\in\{0,1\},
j\in\{1,..,H_0\times W_0\}}
%S^i = \{K^i_j \otimes T_f, j\in [1, p^i]\}
%S^i(j, x, y) = (K^i_j)^T I^t_{x,y}, i\in{0,1}
\label{eq:correlation}
\end{equation}

\begin{figure}
\begin{center}
\vspace{-5mm}
\includegraphics[width=1\linewidth]{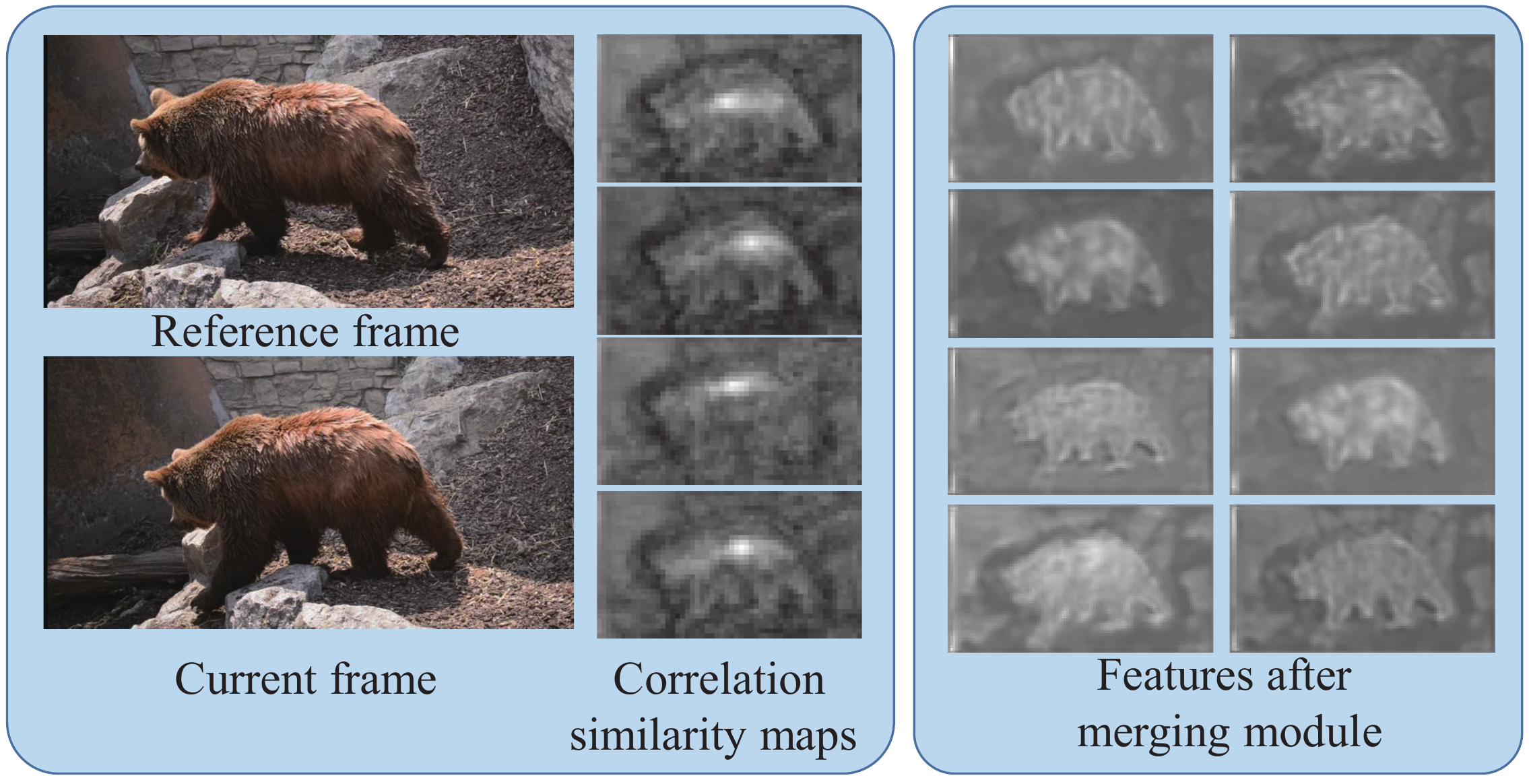}
\end{center}
\vspace{-6mm}
\caption{\textbf{Visualization of the similarity maps}.\ \textsl{Left}:\ the template and current frames, and 4 foreground correlation similarity maps.\ \textsl{Right}:\ the similarity maps after merging.}
\vspace{-5mm}
\label{fig:visual}
\end{figure}

%!!!!!!!!!!!!!!!!!!!!!!!!!!
%Denote $\mathbf{S}$ the tensor of similarity maps in $\mathcal{S}=\mathcal{S}^1\cup\mathcal{S}^0$, we have $\mathbf{S}\in\mathbb{R}^{H_0W_0\times H \times W}$, where $H_0W_0$ is its channel size.\ 
In Fig.~\ref{fig:visual}, we present some examples of the similarity maps.\ Each similarity map is associated with a certain pixel in template frame, whose new position in the current frame is at the maximum (\ie, brightest point) of the similarity map.\ Additionally, in contrast with SiamFC~\cite{siamfc}, since we obtain these maps in a weakly-supervised manner, the contours of the bear, which are essentially preserved for segmentation, are maintained.\ On the right side of Fig.~\ref{fig:visual}, we show some output features of the merging module.\ The object can be distinguished after the merging networks.
%In contrast, SiamFC uses a binary mask, which only maintains location information by highlighting the new position of an object, as the ground truth. 

%, where $(x, y)$ represents the coordinates of each pixel.
%, where "$\otimes$" is the correlation operation.

% down-sampled template mask
%Thus, the similarity maps $S$ can be represented as $S=K*I^t$
%Each reference set contains features of the foreground or background pixels in feature space, whose sizes are $c \times 1 \times 1$, where $c$ is the channel size. 
%The number of features in each set is equal to the pixel number of foreground or background.
%where $p$ is equal to the area of the mask $m^i$. Hence, each $K^i$ has a different number of filters since the proportions of each area are different. 

%$\mathcal{K}^1$ and $\mathcal{K}^0$ are of different cardinalities.\
%For instance, Hu \etal~\cite{videomatch} assemble all the features into one image by calculating their average value. However, this causes a huge decrease in the information. 
\noindent
\textbf{Ranking Attention Module (RAM)}.\
We first utilize the mask of the first frame to filter FG and BG similarity maps.\ Then we design a FG path and a BG path network to process the similarity features.\ Since the number of the FG or BG pixels varies in different videos, the number of FG or BG similarity maps changes dynamically.\ However, regular CNNs require input features with a fixed number of channels.\ 
%template frames, resulting in irregular pixel-level similarity maps for the object and background.\ However, mainstream networks are often designed with regular channels in convolutional layers, requiring regular feature maps as input.\
%The discrepancy between irregularity of FG\ /\ BG similarity maps and regularity of networks is a major problem in our framework.\ 
To tackle this issue, we propose a Ranking Attention Module (RAM) to rank and select important features.\ That is, we learn a scoring scheme for the similarity maps, and then rank and select these maps according to their scores.\ 
%and simply adding more channels is meaningless.\
% since the irregularity of similarity maps is independent from the network
%The RAM module is implemented through a ``Rank-and-Select'' mechanism.\
%into regular maps in a learnable way for our network.

%By this way, the network can learn an attention vector from $\mathbf{S}^{\top}$, and consider the spatial information features of the template 
%and then reshape it as channel-wise attention for similarity $S$. 
%learns a score vector for the similarity maps $\{\mathbf{S}_j^i\}_j\in(0, HW)$.\ Thus, the RAM module learns channel-wise ranking attention score for similarity maps $S \in R^{H_0W_0 \times H \times W}$.\ The reasonable behind it is to transfer channel attention to a special attention.\

As shown in Fig.~\ref{fig:our}, there are three steps in our RAM.\ In the first step, we filter FG (or BG) similarity maps using the mask of the first frame.\ Specifically, we swap the spatial and channel dimensions of similarity maps (reshape $\mathbf{S}\in\mathbb{R}^{H_0W_0\times H\times W}$ into $\mathbf{\hat{S}}\in\mathbb{R}^{HW\times H_0\times W_0}$) and then multiply them with the FG or BG mask (resized to $W_0\times H_0$), respectively.\ Thus, we obtain the FG (or BG) features $\mathbf{\hat{S}}^1$ (or $\mathbf{\hat{S}}^0$).\ In FG component, the features of BG pixels are set as zero, and vice versa.\ 
%The feature tensor $\mathbf{S}^{\top}$ contains the correlation features of the template and current frame in the template space.\ 
%Thus, We transfer channel-wise attention learning to a spacial attention learning problem.
%The correlation operator generate $W_0H_0$ similarity maps 
%At the same time, to select object (background) similarity maps, we  with the feature tensor $\mathbf{\hat{S}}$ pixel-wisely to separate it into a foreground (background) 
%Note that $\mathbf{\hat{S}}^1$ and $\mathbf{\hat{S}}^0$ are of the same size with $\mathbf{\hat{S}}$. 
In the second step, for each similarity map $S_j$, we learn a ranking score $r_j$ which show the importance of each map.\ Taking the FG tensor $\mathbf{\hat{S}}^1$ for instance, to calculate the ranking scores of similarity maps in $\mathbf{\hat{S}}^1$, we use a two-layer network $f_{n}$ strengthened by summing with the channel-wise global max-pooling $f_{max}$ of the tensor $\mathbf{\hat{S}}^1$ in an element-wise manner.\ Larger score indicates greater importance of the corresponding similarity map in $\mathbf{\hat{S}}^1$.\ The channel-wise maximum of each similarity map represents the possibility of corresponding pixel in template frame to find a matching pixel in current frame.\ We define the final FG ranking score metric $\mathbf{R}^1\in\mathbb{R}^{W_0\times H_0}$ as
\begin{equation}
\mathbf{R}^1=f_{n}(\mathbf{\hat{S}}^1)
+
f_{max}(\mathbf{\hat{S}}^1)
.
\end{equation}
Then we reshape $\mathbf{R}^1$ into a vector $\mathbf{r}^1\in\mathbb{R}^{H_0W_0}$.\ Similarly, we can obtain the BG ranking score vector $\mathbf{r}^0$. 

Finally, we rank the similarity maps in $\mathbf{S}^1$ according to the corresponding scores in $\mathbf{r}^1$ from largest to smallest:
\begin{equation}
%\textbf{(Ranking Attention)}:
%\ 
\overline{\mathbf{S}}^1=\text{Rank}(\mathbf{S}^1|\mathbf{r}^1)
.
\end{equation}
% global max-pooling
%neural network
%with $3\times3$ kernels 
If the number of the FG similarity maps $\overline{\mathbf{S}}^1$ is less than the target channel size (set as 256), we pad the ranked feature with zero maps; and if the number is larger than the target channel size, the redundant features are discarded, such that the channel size can be fixed.\ The BG tensor $\mathbf{\hat{S}}^0$ are similarly processed.\ An illustration of the proposed ranking mechanism is shown in Fig.~\ref{fig:rank}.\

%According to Eq.~\ref{eq:S}, since $S^i_j$ is calculated from $K^i_j$, the attention net learns weights for each pixel-level feature $K^i_j$ in the reference frame. 
%Thus, the score of ranking attention can be represented in two ways, the channel-wise attention for current frame $t$'s similarity maps $S \in R^{H_0W_0 \times H \times W}$, or the spatial attention for reference feature $I^r \in R^{C \times H_0 \times W_0}$. 
%$p^i_j=max(S^i_j)$, and the ranking score vector $p$ is Eq.~\ref{eq:p}. We calculate this score via channel wise global max-pooling, as shown in figure~\ref{fig:our}

%\begin{equation}
%p^i=max(S^i)
%\label{eq:p}
%\end{equation}

%%--------MultiVOS--------%% 
\subsection{Extension for Multi-object VOS}
%%%%%%%%%%%%%%%%%%%%%%%%%%%
\label{sec:mifg}

A trivial extension of single-object VOS methods to perform multi-object VOS is to deal with the multiple objects in videos one-by-one.\ But this strategy would be inefficient when there are many objects.\ To make the proposed RANet efficient for multi-object VOS, we share the features extracted by the encoder and also the similarity maps $\mathbf{S}$ computed by correlation for all the $N$ objects.\ Then, for each object $i$ ($i=1,...,N$), we generate its FG and the corresponding BG masks, and segment the FG (or BG) independently using the light-weight decoder.\ Finally, we use a softmax function to compute the final results on VOS.

\begin{figure}
\begin{center}
\includegraphics[width=1\linewidth]{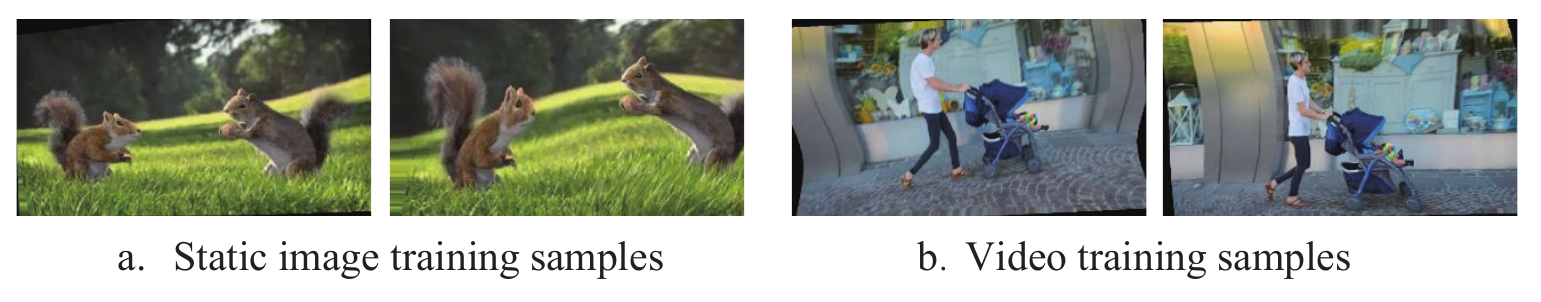}
\end{center}
\vspace{-7mm}
\caption{\textbf{Illustrations of the training samples}. }
%Left: random selected video frames. Right: generated static images pairs.
\label{fig:tps}
\vspace{-6mm}
\end{figure}

%%--------Implement--------%% 
\subsection{Implementation Details}
\label{sec:imple}
Here, we briefly describe the encoder and decoder, and present the detailed network structure in \textsl{Supplemental File}.

\noindent
\textbf{Encoder}.\ The backbone of the two-stream Siamese encoder~\cite{siamfc} is the ResNet-101 network~\cite{resnet}, pre-trained on ImageNet~\cite{imagenet}.\ We replace the batch normalization~\cite{bn2015} with instance normalization~\cite{in2016}.\ The features from the last three blocks are extracted as multi-scale features.\ We reduce the channel sizes of these multi-scale features by four-fold via convolutional layers.\ The features are also resized into the conformable size.\ The $\ell_{2}$ channel-wise normalization~\cite{hoffer2018norm} is added after each convolutional layer for feature pruning and multi-scale merging. 

%We then concatenate the reduced features and pass them through a convolutional layer for multi-scale feature merging in the decoder.
\noindent
\textbf{Decoder}.\ The decoder is a three-level pyramid-like network with skip connection.\
The multi-scale features of current frame extracted by encoder are fed into the decoder.\
However, using all the features in the decoder would bring huge computational costs.\ To speed up our RANet, we first reduce the channel sizes of the multi-scale features using convolutional layers, and then feed them into the decoder.\ 
%The decoder is a three-level pyramid-like network.\ Each level consists of three branches of convolutional layers with dilated sizes of $1$, $6$, and $9$.\ 

% a multi-scale block including
%As can be seen in Fig.~\ref{fig:visual}, 
%Each level has a Multi-scale block, two Conv-block and a Res-block as is shown in Fig.~\ref{fig:merge}. 
%The Multi-scale block has three branches of convolutional layers with dilated sizes of 1, 6, and 9. All the branches are merged through element-wise summation.

%We choose ResNet-101~\cite{resnet} as our backbone, but our method is agnostic as to specific architecture selected.\ 

%%--------Train--------%% 
\subsection{Network Training}
\label{sec:train}
We train our network using the Adam~\cite{adam} with an initial learning rate of $10^{-5}$, to optimize a binary cross-entropy loss.\ During training and test, the input image is resized into $480\times864$.\ We use random Thin Plate Splines (TPS) transformations, rotations ($-30^{\circ}$$\sim$$30^{\circ}$), scaling ($0.75$$\sim$$1.25$), and random cropping for data augmentation, just as~\cite{masktrack}.\ The random TPS transformations are performed by setting $16$ control points and randomly shifting the points within a $15\%$ margin of the image size. 
%\begin{figure}
%\begin{center}
%\includegraphics[width=1\linewidth]{F5/refine.pdf}
%\end{center}
%\vspace{-3mm}
%\caption{The structure of the pyramid network. In each pyramid, there are two Conv\_Blocks, one Res\_block(shown in Fig.~\ref{fig:merge}) and a Multi-scale block(MS\_Block).}
%\label{fig:refine}
%\vspace{-2mm}
%\end{figure}

%\noindent
%\textbf{Cross-correlation.} The purpose of the cross-correlation layer is to locate an object by calculating the pixel-level similarity between the reference filters and the current frame features. The similarity information is arranged as analyzable features for subsequent networks.

%Inspired by the SiamFC network~\cite{siamfc}, we use cross-correlation to measure the similarity between reference filters and target frame features. Our cross-correlation layer is based on a dense correlation layer, which is used in optical flow~\cite{flownet} and geometric matching~\cite{gm}. However, we propose a novel cross-correlation layer for localization and feature arrangement.
%Details of our correlation tracker can be found in Sec.~\ref{Correlation tracker}.

\noindent
\textbf{Pre-train on static images}.\
Following~\cite{masktrack}, we pre-train the proposed RANet using static images.\ To train our RANet for single-object VOS, we use the images from the MSRA10K~\cite{msra10k}, ECSSD~\cite{ecssd}, and HKU-IS~\cite{hku} datasets in the saliency community~\cite{fan2019rethinking,Zhao2019RgbdSal,Fan2019VideoSal,Zhao2019ebd,un03,Lu_2019_CVPR}.\ To train RANet for multi-object VOS, we add the SOC~\cite{soc} and ILSO~\cite{ilso} datasets containing multi-object images.\ %We then employ random TPS transformations, rotations ($-30^{\circ}$$\sim$$30^{\circ}$), scaling ($0.75$$\sim$$1.25$), and cropping to generate more training images.\ We use both all these strategies with the same probability to generate training samples.\ 
Fig.~\ref{fig:tps} (a) shows a pair of generated static images.\ As will be shown in \S\ref{sec:compare} and \S\ref{sec:ablation}, the proposed RANet achieves competitive results when been trained only with static images.

%\caption{Evaluation on DAVIS2017 val and DAVIS2017-testdev datasets. The upper group includes online learning based methods, and the lower group consist of offline methods.}

%frame selection
\noindent
\textbf{Video fine-tuning}.\
Though our RANet can achieve satisfactory results when been trained only with static images, we further exploit its performance by performing video fine-tuning on benchmark datasets.\ To fine-tune our RANet for specific single-object VOS task, we then fine-tune the network on the training set of the DAVIS$_{16}$ dataset~\cite{davis2016}.\ During training, we randomly select two frames with data transformations from one video as the template and current frames, and randomly select the mask of a frame near the current frame (we set the maximum interval as 5).\ We fine-tune our RANet for specific multi-object VOS task on the training set of the DAVIS$_{17}$ dataset~\cite{davis2017}.\ Fig.~\ref{fig:tps} (b) shows an example of paired video training images.

%
% \bf{DAVIS16-val:} Comparison with state-of-the-art methods. The online learning based methods (OSVOS-S, OnAVOS, OSVOS, CINM, MSK) are colored in blue, and the offline methods (RANet, RGMP, FAVOS, PML, SFL, OSMN, CTN, VPN, PLM) are colored in red. And the models run within one second are shown in italics 
% Table generated by Excel2LaTeX from sheet 'Sheet2'

% Table generated by Excel2LaTeX from sheet 'Sheet2'

% Including online learning based methods (OSVOS-S~\cite{osvos-s}, OnAVOS~\cite{onavos}, OSVOS~\cite{osvos}, CINM~\cite{cinm}, MSK~\cite{masktrack} and our RANet with online learning) 
%offline methods (our RANet, RGMP~\cite{rgmp}, FAVOS~\cite{favos}, Videomatch~\cite{videomatch}, PML~\cite{pml}, SFL~\cite{sfl}, OSMN~\cite{osmn}, CTN~\cite{ctn}, VPN~\cite{vpn} and PLM~\cite{plm}).
\begin{table*}[htbp]
\vspace{-2mm}
\centering
%\textcolor[rgb]{ 0,  .439,  .753}{}
\begin{tabular}{r||c|r|c|ccr|ccr}
\Xhline{1pt}
%\rowcolor[rgb]{ .851,  .851,  .851}
\rowcolor[rgb]{ .873,  .91,  0.95}
\multicolumn{1}{c||}{Method} & \multicolumn{1}{l|}{OL} & Time & $\mathcal{J}$\&$\mathcal{F}$$\uparrow$ & $\mathcal{J}$ Mean$\uparrow$ & $\mathcal{J}$ Recall$\uparrow$  & $\mathcal{J}$ Decay$\downarrow$  & $\mathcal{F}$ Mean$\uparrow$  & $\mathcal{F}$ Recall$\uparrow$  & $\mathcal{F}$ Decay$\downarrow$
\\
\hline
\hline
OSVOS\ \ ~\cite{osvos} & \cmark
& 5000 & 80.2 & 79.8 & 93.6 & 14.9 & 80.6 & 92.6 & 15.0
\\
\rowcolor[rgb]{ .94,  .94,  .94}
MaskTrack~\cite{masktrack}  & \cmark
& 12000 & 77.6 & 79.7 & 93.1 & 8.9  & 75.4 & 87.1 & 9.0
\\
CINM\ \ ~\cite{cinm} & \cmark
& 70000 & 84.2 & 83.4 & 94.9 & 12.3 & 85.0 & 92.1 & 14.7  
\\
\rowcolor[rgb]{ .94,  .94,  .94}
OSVOS-S~\cite{osvos-s} & \cmark
& 4500 & 86.6 & 85.6 & 96.8 & 5.5 & 87.5 & 95.9 & 8.2 
\\
OnAVOS~\cite{onavos} & \cmark
& 13000 & 85.5 & 86.1 & 96.1 & \textbf{5.2} & 84.9 & 89.7 & \textbf{5.8} 
\\
\rowcolor[rgb]{ .94,  .94,  .94}
PReMVOS~\cite{premvos} & \cmark
& 38000 & 86.8 & 84.9 & 96.1 & 8.8 & \textbf{88.6} & 94.7 & 9.8 
\\
%\hline
\hline
%\textbf{RANet@0.3s} & \cmark
%& \textbf{0.30} & 86.3 & 86.1 & 97.2 & 6.7 & 86.5 & 96.3 & 6.3 
%\\
%\textbf{RANet@1s} & \cmark
%& 1.00 & 86.7 & 86.4 & \textbf{97.4} & 6.7 & 86.9 & 96.1 & 6.9 
%\\
%\textbf{RANet@2s} & \cmark 
%& 2.00 & 86.8 & 86.6 & \textbf{97.4} & 6.5 & 87.0 & 96.2 & 6.6 
%\\
RANet+ & \cmark
& 4000 &
\textbf{87.1} & \textbf{86.6} & \textbf{97.0} & 7.4 & 87.6 & \textbf{96.1} & 8.2
\\
\hline
\hline
\rowcolor[rgb]{ .94,  .94,  .94}
PLM~\cite{plm}  & \xmark
& 500 & 66.4 & 70.2 & 86.3 & 11.2 & 62.5 & 73.2 & 14.7
\\
VPN~\cite{vpn}  & \xmark     
& 630 & 67.9 & 70.2 & 82.3 & 12.4 & 65.5 & 69.0 & 14.4 
\\
\rowcolor[rgb]{ .94,  .94,  .94}
SiamMask~\cite{siammask} & \xmark   
& \textbf{28} & 70.0 & 71.7 & 86.8 & \textbf{3.0} & 67.8 & 79.8 & \textbf{2.1}  
\\
CTN~\cite{ctn}  & \xmark    
& 30000 & 71.4 & 73.5 & 87.4 & 15.6 & 69.3 & 79.6 & 12.9 
\\
\rowcolor[rgb]{ .94,  .94,  .94}
OSMN~\cite{osmn} & \xmark    
& 130 & 73.5 & 74.0 & 87.6 & 9.0 & 72.9 & 84.0 & 10.6
\\
SFL~\cite{sfl}  & \xmark     
& 7900 & 76.1 & 76.1 & 90.6 & 12.1 & 76.0   & 85.5 & 10.4
\\
\rowcolor[rgb]{ .94,  .94,  .94}
PML\ \ ~\cite{pml}  & \xmark    
& 280 & 77.4 & 75.5 & 89.6 & 8.5 & 79.3 & 93.4 & 7.8  
\\
VideoMatch~\cite{videomatch} & \xmark  
& 320    & -    & 81.0   & -    & -    & -    & -    & -   
\\
\rowcolor[rgb]{ .94,  .94,  .94}
FAVOS\ \ ~\cite{favos} & \xmark   
& 1800 & 81.0 & 82.4 & 96.5 & 4.5 & 79.5 & 89.4 & 5.5  
\\
FEELVOS~\cite{Voigtlaender2019FEELVOS} & \xmark  
& 510 & 81.7 & 81.1 & 90.5 & 13.7 & 82.2 & 86.6 & 14.1 
\\
\rowcolor[rgb]{ .94,  .94,  .94}

RGMP~\cite{rgmp} & \xmark  
& 130 & 81.8 & 81.5 & 91.7 & 10.9 & 82.0 & 90.8 & 10.1 
\\
%\hline
\hline
%\rowcolor[rgb]{ .94,  .94,  .94}
RANet & \xmark  
& \textbf{33} & \textbf{85.5} & \textbf{85.5} & \textbf{97.2} & 6.2 & \textbf{85.4} & \textbf{94.9} & 5.1 
\\
\hline
\end{tabular}
\vspace{-3mm}
\caption{\textbf{Comparison on objective metrics and running time (in milliseconds) by different methods on the DAVIS$_{16}$-val dataset}.\ The best results of online learning (OL) based methods and offline methods are both highlighted in bold.}
\label{tab:16val}%
\vspace{-5mm}
\end{table*}%

%%%%%%%%%%%%%%

% Table generated by Excel2LaTeX from sheet 'Sheet4'
\begin{table}[t]
\vspace{1.5mm}
\centering
\begin{tabular}{r||ccr}
\Xhline{1pt}
\rowcolor[rgb]{ .873,  .91,  0.95}
\multicolumn{1}{c||}{Method} & $\mathcal{J}$ Mean$\uparrow$& $\mathcal{J}$ Recall$\uparrow$& $\mathcal{J}$ Decay$\downarrow$
\\
\hline
\hline
BVS~\cite{bvs}  & 66.5 & 76.4 & 26.0
\\
\rowcolor[rgb]{ .94,  .94,  .94}
OFL~\cite{ofl}  & 71.1 & 80.0   & 22.7
\\
VPN~\cite{vpn}  & 75.0   & 90.1 & 9.3 
\\
\rowcolor[rgb]{ .94,  .94,  .94}
CTN~\cite{ctn}  & 75.5 & 89.0   & 14.4 
\\
MaskTrack~\cite{masktrack}  & 80.3 & 93.5 & 8.9
\\
%MaskTrack-OL~\cite{masktrack} & 69.9 & -   & - 
%\\
\hline
\hline
\rowcolor[rgb]{ .94,  .94,  .94}
RANet & 83.2 & 94.2 & 9.3
\\
RANet+OL & \textbf{86.2} & \textbf{96.2} & \textbf{7.6}
\\
\hline
\end{tabular}%
\vspace{-3mm}
\caption{\textbf{Comparison of different methods without video fine-tuning on DAVIS$_{16}$-trainval dataset}.\ ``RANet+OL'' denotes the proposed RANet boosted by OL techniques.}
\label{tab:16trainval}%
\vspace{-5mm}
\end{table}%

%%--------EXperiments--------%% 
\section{Experiments}
\label{sec:exp}
In this section, we first describe our experimental protocol (\S\ref{sec:protocol}), and then compare the proposed ranking attention network (RANet) with the state-of-the-art VOS methods (\S\ref{sec:compare}).\ We next perform a comprehensive ablation study to gain deeper insights into the proposed RANet, especially on the effectiveness of the ranking attention module (\S\ref{sec:ablation}).\ Finally, we present the visual results to show the robustness of RANet against challenging scenarios (\S\ref{sec:visual}).\ More results are provided in the \emph{Supplementary File}.

%introduce the implementation details in Sec.~\ref{details}, results comparing our method to state-of-the-art techniques in Sec.~\ref{Comparison}, evaluation of each component of our method in Sec.~\ref{Ablation}, and feature visualization in Sec.~\ref{visualization}
\subsection{Experimental Protocol}
\label{sec:protocol}

\noindent
\textbf{Training datasets}.\ We evaluate the proposed RANet on the DAVIS$_{16}$~\cite{davis2016} and DAVIS$_{17}$~\cite{davis2017} datasets.\ The DAVIS$_{16}$ dataset~\cite{davis2016} contains 50 videos ($480$p), annotated with pixel-level object masks (one per sequence) densely on the total 3455 frames, and it is divided into a training set ($30$ videos), and a validation set ($20$ videos).\ The DAVIS$_{17}$ dataset~\cite{davis2017}, that contains videos with multiple objects, is an extension of DAVIS$_{16}$, and it contains a training set with $60$ videos, a validation set with $30$ videos, and a test-dev set with $30$ videos.\ In all datasets, there is no overlap among the training, validation, and test sets.
%, DAVIS16~\cite{davis2016}, DAVIS17~\cite{davis2017}, Segtrack~\cite{segtrack}, Jumpcut~\cite{jumpcut}. 
%Youtube-objects~\cite{}

\noindent
\textbf{Testing phase}.\
%To segment a video into objects and background, 
Similar to SiamFC~\cite{siamfc}, we crop the first frame and extract the features as the template features ($\mathcal{K}$ in \S\ref{sec:ccsim}), then compute the similarity maps between the features of template frame and of the test frames one-by-one, and finally segment the current test frame.\ The video data used are in different goals: 1) to evaluate our RANet for single-object VOS, we test it on the validation set ($20$ videos) of~\cite{davis2016}; 2) to judge the effectiveness of our RANet trained only on static images, we evaluate it on the $50$ videos of the whole DAVIS$_{16}$ dataset; 3)
to assess our RANet for multi-object VOS, we evaluate it on the validation and test sets of~\cite{davis2017}, which respectively contain $30$ videos. 
%\noindent
%\textbf{Online learning}.\ 
To compare with OL based methods, we follow ~\cite{osvos,masktrack}, fine-tuning on the first frame with data augmentation for each video. We use the same training strategy as pre-training on static images, but the learning rate is $10^{-6}$.

%Inspired by single-object tracking methods~\cite{cfnet,siammask}, we employ the center crop strategy to search the object area in the test frame, according to its bounding box in the previous frame. We set the searching area to be 1.5 times larger than the minimum enclosing rectangle of the previous frame's segmentation, then crop and zoom-in the test frame image without changing the aspect ratio.

\noindent
\textbf{Evaluation metrics}.\ We use seven standard metrics suggested by~\cite{davis2016}: three region similarity metrics $\mathcal{J}$ Mean, $\mathcal{J}$ Recall, $\mathcal{J}$ Decay; three boundary accuracy metrics $\mathcal{F}$ Mean, $\mathcal{F}$ Recall, $\mathcal{F}$ Decay; and $\mathcal{J}\&\mathcal{F}$ Mean, which is the average of $\mathcal{J}$ Mean and $\mathcal{F}$ Mean.

%%--------Comparison--------%% 
\subsection{Comparison to the state of the art}
\label{sec:compare}

\noindent
\textbf{Comparison Methods}.\ For single object VOS, we compare our RANet with $6$ state-of-the-art OL based and $11$ offline methods~\cite{osvos-s,masktrack,osvos,cinm,onavos,premvos,plm,vpn,siammask,ctn,osmn,sfl,pml,videomatch,favos,Voigtlaender2019FEELVOS,rgmp} in Table~\ref{tab:16val}, including OSVOS-S~\cite{osvos-s}, PReMVOS~\cite{premvos}, RGMP~\cite{rgmp}, FEELVOS~\cite{Voigtlaender2019FEELVOS}, \textit{etc}.\ To evaluate our RANet trained with static images, we compare it with some methods~\cite{bvs,ofl,vpn,ctn,masktrack} without using DAVIS training set.\ For multi-object VOS, we compare with some state-of-the-art offline methods~\cite{osvos,onavos,favos,osmn,videomatch}, and also list results of some OL based methods~\cite{cinm,osvos-s,onavos,osvos,videomatch} for reference.

\noindent
\textbf{Results on DAVIS$_{16}$-val}.\ As shown in Table~\ref{tab:16val}, without online learning (OL) technique, our RANet still achieves a $\mathcal{J}\&\mathcal{F}$ Mean of $85.5\%$ at a speed of $33$ milliseconds ($30$FPS).\ For RANet, its metric results are higher than all the methods without OL techniques, while its speed is higher than all the compared methods, except SiamMask~\cite{siammask}.\ But please note that SiamMask performs badly on objective metrics, e.g., $70.0\%$ at $\mathcal{J}\&\mathcal{F}$, $15.5$ points lower than our RANet.\ Even when compared with the state-of-the-art OL based methods such as OSVOS-S~\cite{osvos-s} and OnAVOS~\cite{onavos}, our offline RANet achieves comparative results.\ The RANet can be improved by OL techniques.\ The OL boosted RANet, denoted as RANet+, achieves a $\mathcal{J}\&\mathcal{F}$ Mean of $87.1\%$, outperforming all OL based VOS methods.  

%according to the runtime, which is related to the number of iterations for online training. Notice that, when the runtime is 1 second, our method outperforms all the state-of-the-art methods. 

% Table generated by Excel2LaTeX from sheet 'Sheet5'
\begin{table}[t]
\vspace{-2mm}
\centering
\scriptsize
\begin{tabular}{r||c|cc|cc}
\Xhline{1pt}
\rowcolor[rgb]{ .873,  .91,  0.95}
&
& 
\multicolumn{2}{c|}{DAVIS$_{17}$-val} 
& 
\multicolumn{2}{c}{DAVIS$_{17}$-testdev} 
\\
%\cline{4-7} 
\rowcolor[rgb]{ .873,  .91,  0.95}
\multicolumn{1}{c||}{\multirow{-2}{*}{Method}}
%\normalsize{Method}
& 
\multirow{-2}{*}{OL}
%& 
%\multirow{-2}{*}{Time}
& $\mathcal{J}$\&$\mathcal{F}\uparrow$  & $\mathcal{J}$ Mean$\uparrow$  & $\mathcal{J}$\&$\mathcal{F}\uparrow$  & $\mathcal{J}$ Mean$\uparrow $
\\
\hline
\hline
CINM\ \ ~\cite{cinm} & \cmark &
\textbf{70.6} & \textbf{67.2} & \textbf{67.5} & \textbf{64.5} 
\\
\rowcolor[rgb]{ .94,  .94,  .94}
OSVOS-S~\cite{osvos-s} & \cmark &
68.0 & 64.7 & 57.5 & 52.9 
\\
OnAVOS~\cite{onavos} & \cmark & 
65.4 & 61.6 & 52.8 & 49.9 
\\
\rowcolor[rgb]{ .94,  .94,  .94}
OSVOS\ \ ~\cite{osvos} & \cmark & 
60.3 & 56.6 & 50.9 & 47.0 
\\
VideoMatch~\cite{videomatch} & \cmark & 
61.4 & -    & -    & - 
\\
\hline
\hline
\rowcolor[rgb]{ .94,  .94,  .94}
OSVOS\ \ ~\cite{osvos} & \xmark & 
36.6 & -    & -    & - 
\\
OnAVOS~\cite{onavos} & \xmark & 
39.5 & -    & -    & - 
\\
\rowcolor[rgb]{ .94,  .94,  .94}
FAVOS\ \ ~\cite{favos} & \xmark & 
58.2 & 54.6 & 43.6 & 42.9 
\\
OSMN~\cite{osvos-s} & \xmark & 
54.8 & 52.5 & 41.3 & 37.7 \\
\rowcolor[rgb]{ .94,  .94,  .94}
VideoMatch~\cite{videomatch} & \xmark & 
56.5 & -    & -    & - 
\\
\hline
RANet & \xmark &
\textbf{65.7} & \textbf{63.2} & \textbf{55.3} & \textbf{53.4}
\\
\hline
\end{tabular}%
\vspace{-3mm}
\caption{\textbf{Comparison of different methods on DAVIS$_{17}$-val and DAVIS$_{17}$-testdev datasets}.\ The methods are divided into two groups according to whether online learning (OL) technique is employed or not.\ }
%``-'' means the authors have not released the results.}
\label{tab:17val}%
\vspace{-4mm}
\end{table}%

\noindent
\textbf{Results on DAVIS$_{16}$-trainval}.\
We also evaluate the performance of our RANet trained only with static images (i.e., without video fine-tuning).\ MaskTrack~\cite{masktrack} has the most similar setting as our RANet in this case, since it also uses only static images to train its networks.\ Contrast to MaskTrack, our RANet does not rely on OL techniques, speeding up for nearly a hundred times faster.\ In Table~\ref{tab:16trainval}, we list the results of different methods that do not require fine-tuning/training on video data.\ Again, our RANet outperforms all the other methods by a clear margin.

% \textbf{DAVIS16-trainval:} Comparison with the state-of-the-art methods without video fine-tuning on the whole DAVIS16 dataset. Online learning disabled methods are denoted with $-$, and "+" represents adding online learning step.

\noindent
\textbf{Results on DAVIS$_{17}$ dataset}.\ The DAVIS$_{17}$ dataset is challenging due to multi-object scenarios.\ To evaluate our RANet on DAVIS$_{17}$-val and DAVIS$_{17}$-test sets, we use the RANet trained on multi-instance static images and the DAVIS$_{17}$-train dataset, as described in \S\ref{sec:train}.\ In Table~\ref{tab:17val}, we show the comparison of our RANet with state-of-the-art VOS methods.\ It can be seen that on the DAVIS$_{17}$-val dataset, our RANet achieves higher metric results than the \emph{w/o} OL methods.\ Furthermore, on the more challenging DAVIS$_{17}$-testdev dataset, our RANet even outperforms the OL based method OnAVOS in terms of $\mathcal{J}$ Mean.

\begin{figure}[t]%[htbp]
\vspace{-4mm}
\begin{center}
\includegraphics[width=1\linewidth]{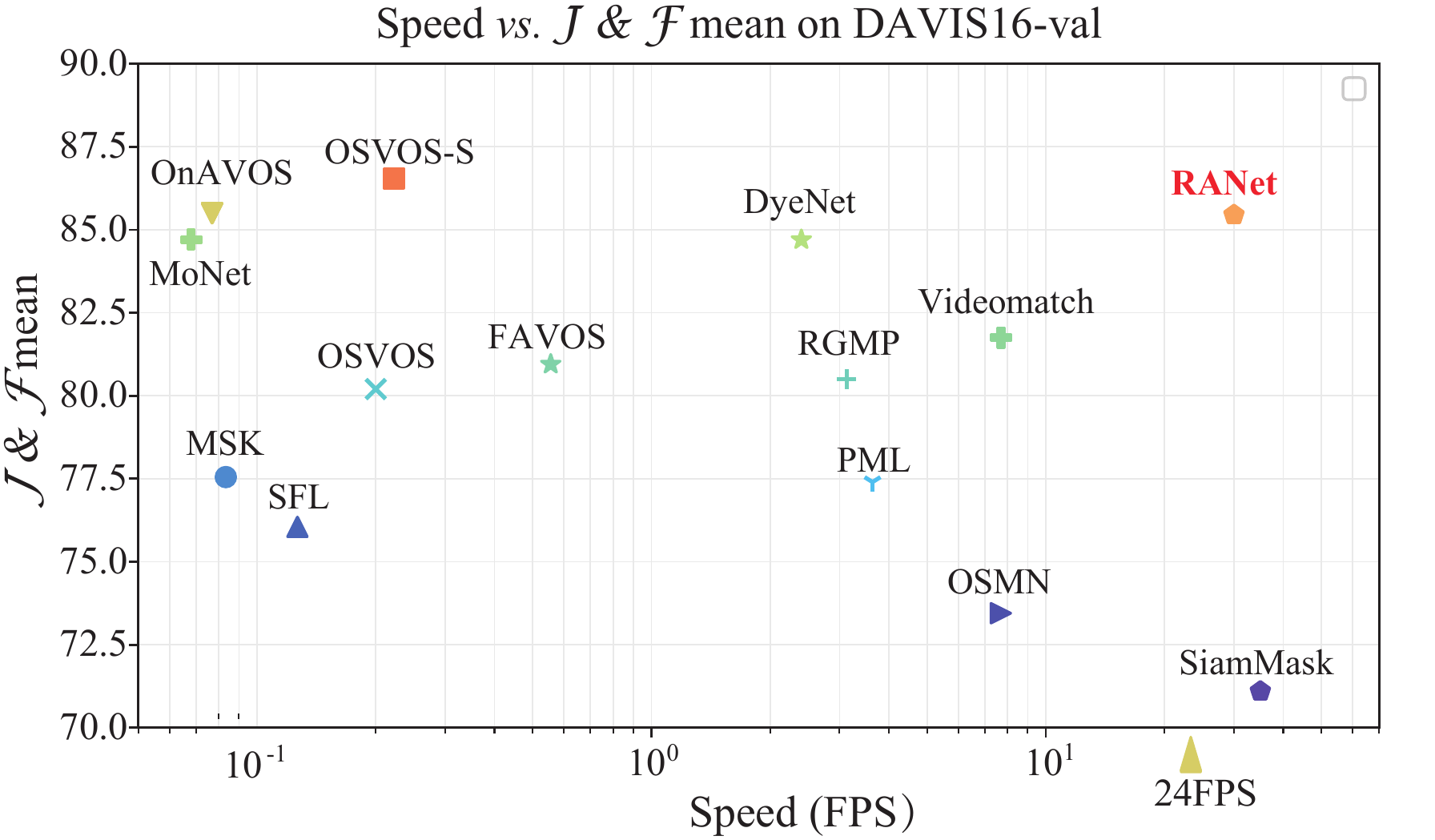}
\end{center}
\vspace{-6mm}
\caption{\textbf{Comparison of $\mathcal{J}\&\mathcal{F}$ Mean and Speed} (in FPS) by different methods on DAVIS$_{16}$-val dataset.}
\vspace{-2mm}
\label{fig:speed}
\end{figure}

\noindent
\textbf{Speed}.\
Here, we evaluate the speed-accuracy performance of different methods on DAVIS$_{16}$-val set.\ Our RANet runs on a TITAN Xp GPU.\ In Table~\ref{tab:16val}, we list the average time of different methods processing a frame of 480p resolution.\ Note that the proposed RANet spends 33 milliseconds on each frame, much faster than most of the previous methods.\ As shown in Fig.~\ref{fig:speed}.\ The recently proposed method SiamMask~\cite{siammask} is a little faster than our RANet but at expenses of much lower results on $\mathcal{J}\&\mathcal{F}$ Mean than ours.\ 
% On DAVIS$_{17}$-val dataset, the speed of RANet is $70$ milliseconds per frame.

%\textbf{Segtrack dataset:}
%We present the evaluation results on the Segtrack dataset in Table~\ref{}. Our method outperforms most of the methods even though our network is not fine-tuned.

%\noindent
%\textbf{Jumpcut dataset:}
%We present the evaluation results on the Jump-Cut dataset in Table~\ref{}. We follow the evaluation in~\cite{jumpcut} and compute the error rates of different methods. The transfer distance d is equal to 16. our method outperforms the baselines on this dataset with an average error rate that is *** lower than the best competing baseline.

%%--------Ablation--------%% 
\subsection{Validation of the Proposed RANet}
\label{sec:ablation}

We now conduct a more detailed examination of our proposed RANet on the VOS task.\ We assess 1) the contribution of the proposed ranking attention module (RAM) to RANet; 2) the importance of correlation layer (CL) to RANet; 3) the influence of propagating previous frame's mask (PM) on RANet; 4) the effect of static image pre-train (IP) and video fine-tuning (VF) on RANet; and 5) the impact of online learning (OL) technique to RANet.

% We then multiply the foreground (background) mask with the feature tensor $\mathbf{S}^{\top}$ pixel-wisely to separate it into a foreground feature tensor $\mathbf{\hat{S}}^1$ and a background feature tensor $\mathbf{\hat{S}}^0$, respectively.\ In the foreground component, the features of background pixels are set as zero, and vice versa.\ Note that $\mathbf{\hat{S}}^1$ and $\mathbf{\hat{S}}^0$ are of the same size with $\mathbf{S}^{\top}$.
\begin{table}[t]
  \centering
    \begin{tabular}{c||ccc}
    \Xhline{1pt}
    \rowcolor[rgb]{ .873,  .91,  0.95} 
Variant
    & \multicolumn{1}{l}{\emph{w/ RAM}} 
    & \multicolumn{1}{l}{\emph{w/o Ranking}}
    & \multicolumn{1}{l}{\emph{Maximun}} 
\\
    \hline
%\multirow{2}[2]{*}{J Mean}
$\mathcal{J}$ Mean & 85.5 &   81.9   & 81.1 
\\
% & 79.9 & 77.1 & 78.9 \\
    \hline
    \end{tabular}%
    \vspace{-3mm}
    \caption{\textbf{Comparison of $\mathcal{J}$ Mean by different variants of RANet on DAVIS$_{16}$-val dataset}.\ 
    %The original RANet is \emph{w/ RAM}, while two other baselines are \emph{w/o Ranking}, and \emph{Maximum}. 
    }
    \vspace{-4mm}
  \label{tab:ram}%
\end{table}%

\noindent
\textbf{1.\ Does the proposed ranking attention module contribute to RANet}?\
To evaluate the contribution of the proposed RAM module to RANet on VOS task, we compare the original RANet, we call it \emph{w/ RAM}, with two baselines.\ For the first one, \emph{w/o Ranking}, we maintain all the similarity maps in $\mathcal{S}$, and obtain FG (or BG) similarity maps $\mathbf{S}^1$ (or $\mathbf{S}^0$)$\in$$\mathbb{R}^{H_0W_0\times H\times W}$ by setting corresponding BG (or FG) as zeros according to the template mask.\ For the second one, \emph{Maximum}, instead of using RAM to obtain abundant embedding maps, we employ channel-wise maximum operation, which is also used in~\cite{Voigtlaender2019FEELVOS}, on the similarity maps $\mathbf{S}^1$ and $\mathbf{S}^0$, respectively, to get one FG and one BG map $\mathbf{S}^1_M,\mathbf{S}^0_M\in\mathbb{R}^{H\times W}$.\ Then we feed them into the decoder.\ %We called it:.\ %\emph{Maximum} calculates the maximum value of the similarity maps at each pixel. 
%by setting the BG/FG pixels as zeros, and then feed them into the decoder.
%Thus the position distribution of the FG\ /\ BG similarity maps changes along with the distribution of the FG\ /\ BG pixels of the template masks.
% Table generated by Excel2LaTeX from sheet 'ab'

The comparison of RANet \emph{w/ RAM}, \emph{w/o Ranking}, and \emph{Maximum} is listed in Table~\ref{tab:ram}.\ It can be seen that, the RANet \emph{w/ RAM} achieves $3.6\%$ and $4.4\%$ higher than the baselines \emph{w/o Ranking} and \emph{Maximum}, respectively.\ The RANet \emph{w/o Ranking} organizes the similarity maps based on the spacial information of the template frame, while the RANet with \emph{Maximum} losses most useful information in similarity maps by only extracting the maximum values.

\begin{figure*}
\begin{center}
\vspace{-4mm}
\includegraphics[width=0.95\linewidth]{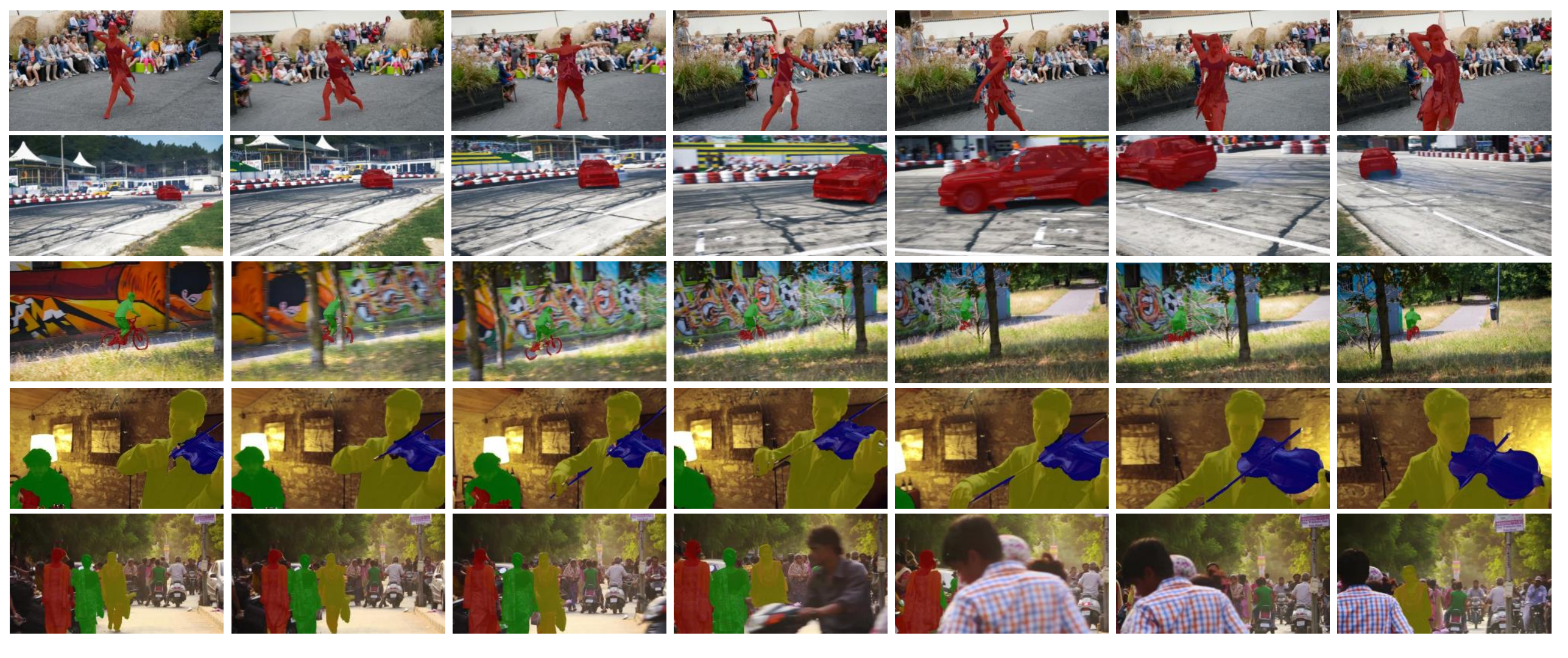}
\end{center}
\vspace{-9mm}
\caption{\textbf{Qualitative results of the proposed RANet on challenging VOS scenarios}.\ The test frames are from videos in the DAVIS$_{16}$ set ($1$-st and $2$-nd rows), the DAVIS$_{17}$-val set ($3$-rd row), and the DAVIS$_{17}$-testdev set ($4$-th and $5$-th rows).}
\label{fig:results}
\vspace{-5mm}
\end{figure*}

\noindent
\textbf{2.\ How important is the correlation and RAM to our RANet}?\ 
To evaluate the importance of correlation layer in our RANet, we remove the correlation layer, and simply concatenate the features extracted by the encoder, as RGMP~\cite{rgmp} does.\ The following RAM module is also meaningless and is removed.\ Thus we have a new variant of RANet: \emph{-CL}.\ However, as shown in Table~\ref{tab:ablation}, the performance of this variant is very bad (67.5\% on $\mathcal{J}$ Mean).\ Thus, the correlation layer is important to our RANet, and serves as the base for the proposed RAM module. 

\begin{table}[t]%htbp
\centering
%\vspace{-7mm}
\vspace{2mm}
\begin{tabular}{l||c|ccccc}
\Xhline{1pt}
\rowcolor[rgb]{ .873,  .91,  0.95}
\multicolumn{1}{c||}{Method}  
& origin & \emph{-CL}  & \emph{-PM} & \emph{-IP} & \emph{-VF} 
\\
\hline
\hline
RGMP~\cite{rgmp} & 81.5 & - & 73.5 & 68.6 & 55.0
\\
RANet & 85.5 & 67.5 & 81.4 & 73.2 & 79.9
\\
\hline
\end{tabular}
\vspace{-3mm}
\caption{\textbf{Ablation study of RANet on $\mathcal{J}$ Mean}.\ CL, PM, IP, and VF mean Correlation Layer, Previous frame's Mask, static Image Pre-train, and Video Fine-tuning, respectively.}
\label{tab:ablation}%
\vspace{-2mm}
\end{table}%
% center crop, & -CC & - & 77.7 

% Table generated by Excel2LaTeX from sheet 'online'
\begin{table}[t]
\vspace{1mm}
\centering
    \begin{tabular}{c||c|cccc}
    \Xhline{1pt}
\rowcolor[rgb]{ .873,  .91,  0.95}
    Metric & offline & \multicolumn{4}{c}{+online learning}
    \\
    \hline
    \hline
    %RANet w/o VF & 80.9 & 80.7 & 81.4 & 82.4 & 82.4 
    %\\
    $\mathcal{J}$\&$\mathcal{F}$ Mean & 85.5 & 86.2 & 86.8 & 86.9 & 87.1 \\
    Time & 0.033 & 0.30 & 1.00 & 1.50 & 4.00 
    \\
    \hline
    \end{tabular}%
    \vspace{-3mm}
    \caption{\textbf{Influence of online learning to RANet} with different iterations on $\mathcal{J}$\&$\mathcal{F}$ Mean and runtime (in seconds).} 
\label{tab:online}%
\vspace{-4mm}
\end{table}%

\noindent
\textbf{3.\ How does the previous frame's mask (PM) influence our RANet}?\ 
We study how PM influences our RANet.\ To this end, we set all the pixels of the PM as zero, and re-train our RANet.\ Thus we have a baseline of \emph{-PM}.\ Results in Table~\ref{tab:ablation} shows that, the variant \emph{-PM} of RANet will drop $\mathcal{J}$ Mean by $4.1$ points.\ This indicates that the temporal information propagated by PM is very useful for our RANet.

%\noindent
%\textbf{2. How each component influences RANet?}\ Similar to RGPM has similar settings evaluate the influence of the previous frame’s mask (PM), static image pre-train, video fine-tuning (VF) and online learning on RANet. The evaluation is in the same spirit as RGPM~\cite{rgmp}.\ From Table~\ref{tab:ablation}, we can see that each component is important to the proposed RANet. But RANet is more robust on component removal than RGPM. To evaluate the model without the previously predicted mask, we simply set all the pixels of the mask to zero, and use the same settings to re-train our model. Results show that the time-sequential information provided by mask propagation is necessary for our model. 

\noindent
\textbf{4.\ What are the effects of pre-training on static images and video fine-tuning in our RANet}?\
To answer this question, we study how each training strategy affects the performance of RANet.\ We first train RANet only on video data and have a baseline: \emph{-IP}.\ We then train RANet only on static images and have the second baseline: \emph{-VF}.\ The results of $\mathcal{J}$ Mean by the variants \emph{-IP} and \emph{-VF} on DAVIS$_{16}$-val dataset are listed in Table~\ref{tab:ablation}.\ As can be seen, both baselines drop significantly on $\mathcal{J}$ Mean when compared to the original RANet.\ Specifically, static image pre-train (IP) improves the $\mathcal{J}$ Mean from $73.2\%$ to $85.5\%$, while video fine-tuning (VF) improves the $\mathcal{J}$ Mean by $5.6$ points.\ The performance drops (from 85.5\% to 73.2\%) of removing IP is mainly due to the over-fitting of RANet on the DAVIS$_{16}$-training set, which only contains 30 single-object videos.

\noindent
\textbf{5.\ The trade-off between performance and speed using online learning}.\
In Table~\ref{tab:online}, we also show the performance and run-time of RANet with or without OL technique.\ One can see that, as the number of iterations increases in OL, the results of our RANet on $\mathcal{J}\&\mathcal{F}$ Mean are continuously improved with different extents, while at a cost of speed.

%\textbf{5. Is center crop helpful?} There are two advantages of using center crop in segmentation. The first is that if removes the pixels that are far from the target, since we only detect the object around its location in the previous frame. The second advantage is that small objects become larger after cropping, which leads to better segmentation. From Table~\ref{tab:ablation}, we can find center crop improves the performance.

%%--------Visualize--------%% 
\subsection{Qualitative Results}
\label{sec:visual}
In Fig.~\ref{fig:results}, we show some qualitative visual results of the proposed RANet on the DAVIS$_{16}$ and DAVIS$_{17}$ datasets.\ It can be seen that, the RANet is very robust against many challenging scenarios, such as appearance changes ($1$-st row), fast motion ($2$-nd row), occlusions ($3$-th row), and multi-objects ($4$-rd and $5$-th rows), \textit{etc}.\
% We also show some failure cases in Fig.~\ref{fig:failure}. Since our similarity is measured on the pixel-level, it is difficult to distinguish similar instances that are close in the spatial domain.

%The left side of Fig.~\ref{fig:visual} shows the similarity score maps. In each similarity map, the maximum similarity represents the new position of the point in the reference image. Additionally, the contours of the objects are maintained, which is essential for segmentation. In contrast, SiamFC uses a binary mask, which only maintains location information by highlighting the new position of an object as the ground truth. 
%Finally, Videomatch uses the averaged similarity maps as the final segmentation, which encourages all the similarity maps to be more similar to the ground truth.
%On the right side of Fig.~\ref{fig:visual}, Our merging network can better distinguish objects.

%%--------Conclusion--------%% 
\section{Conclusion}

In this work, we proposed a real-time and accurate VOS network, which runs at 30 FPS on a single Titan Xp GPU.\ The proposed ranking attention network (RANet) end-to-end learned the pixel-level feature matching and mask propagation for VOS.\ A ranking attention module was proposed to better utilize the similarity features for fine-grained VOS performance.\ The network treated the point-to-point matching feature as a guidance instead of the final results, to avoid noisy predictions.\ Experiments on DAVIS$_{16}/_{17}$ datasets demonstrate that our RANet achieves state-of-the-art performance on both segmentation accuracy and speed.

This work can be further extended.\ First, the proposed ranking attention module can be applied to other applications such as object tracking~\cite{siammask} and stereo vision~\cite{Khamis2018StereoNet}.\ Second, better propagation~\cite{flownet, flownet2} or local matching~\cite{Voigtlaender2019FEELVOS} techniques can be employed for better VOS performance.

\vspace{2mm}
\noindent
\textbf{Acknowledgements}.\
We thank Dr.\ Song Bai on the initial discussion of this project.

\clearpage
{
%\small
%\footnotesize
\balance
\bigsize
\bibliographystyle{ieee}%ieee
\bibliography{vos}
}

\end{document}